\documentclass[runningheads]{llncs}
\usepackage[T1]{fontenc}
\usepackage{graphicx}
\usepackage{amsfonts,amssymb}
\usepackage{color}
\usepackage{algorithm}
\usepackage[noend]{algpseudocode}

\begin{document}
\title{Synchronous Image-Label Diffusion Probability Model with Application to Stroke Lesion Segmentation on Non-contrast CT}
%
%

\author{Jianhai Zhang$^1$ Tonghua Wan$^2$ Ethan MacDonald$^1$ Bijoy Menon$^1$\\ Aravind Ganesh$^1$ Qiu Wu$^2$}

%
\institute{1. University of Calgary\\2. Huazhong University of Science and Technology\\
}
%

\maketitle              

\negthinspace\negthinspace\negthinspace\negthinspace
\begin{abstract}
\negthinspace Stroke lesion volume is a key radiologic measurement for assessing the prognosis of Acute Ischemic Stroke (AIS) patients, which is challenging to be automatically measured on Non-Contrast CT (NCCT) scans. Recent diffusion probabilistic models have shown potentials of being used for image segmentation. In this paper, a novel Synchronous image-label Diffusion Probability Model (SDPM) is proposed for stroke lesion segmentation on NCCT using Markov diffusion process. The proposed SDPM is fully based on a Latent Variable Model (LVM), offering a complete probabilistic elaboration. An additional net-stream, parallel with a noise prediction stream, is introduced to obtain initial noisy label estimates for efficiently inferring the final labels. By optimizing the specified variational boundaries, the trained model can infer multiple label estimates for reference given the input images with noises. The proposed model was assessed on three stroke lesion datasets including one public and two private datasets. Compared to several U-net and transformer based segmentation methods, our proposed SDPM model is able to achieve state-of-the-art performance. The code is publicly available.

\negthinspace\negthinspace
\end{abstract}

\section{Introduction}
\negthinspace
Stroke lesion volume is a key radiologic measurement in assessing prognosis of Acute Ischemic Stroke (AIS) patients~\cite{bucker2017associations}. Early assessment of patient outcome is beneficial to inform patients about future perspectives as soon as possible, and to enable treating physician to adapt and personalize the treatments and rehabilitation plans.
Ischemic stroke lesions, such as hemorrhagic and ischemic infarct, are typically measured on post treatment Non-Contrast CT (NCCT) scans.
Manual contouring of stroke lesions are still clinically deemed as gold standard for volume measurement even though it is time consuming and observer dependent~\cite{ospel2021radiologic,ospel2021detailed}.
Regardless of many attempts to automate segmentation for stroke lesions~\cite{dobshik2023acute,gauriau2023head,liang2021symmetry}, there are still no methods well-established for NCCT, as cerebral CT is limited due to low signal to noise ratio, low contrast of soft tissues, partial volume effects, and acquisition variability across different scanners~\cite{liu2020attention}.

This study aims to develop a Diffusion Probabilistic Model (DPM) based approach~\cite{sohl2015deep} to accurately segmenting ischemic or hemorrhagic lesion on NCCT.
Recently, the methods~\cite{baranchuk2021label,wu2022medsegdiff,amit2021segdiff} based on DPM have received increasing attention for medical image segmentation.
The attention is warranted because of the powerful denoising mechanism against even high degrees of noise contamination.
In essence, a progressive denoising process~\cite{song2019generative} for image generation (or potential image labels) is foundationally different from the previous techniques outputting results immediately. Moreover, the introduction of denoising for training indeed improves the robustness of the prediction because of the massive observations with different degrees of noises. Thus, the stability of segmentation performance is able to be guaranteed. Nevertheless, the current diffusion models for medical image segmentation~\cite{wu2022medsegdiff,amit2021segdiff} are simply treating the image as the conditional input fed to the models without probabilistic interpretability.

\begin{figure*}[t]
\center
\includegraphics[width=0.70\textwidth]{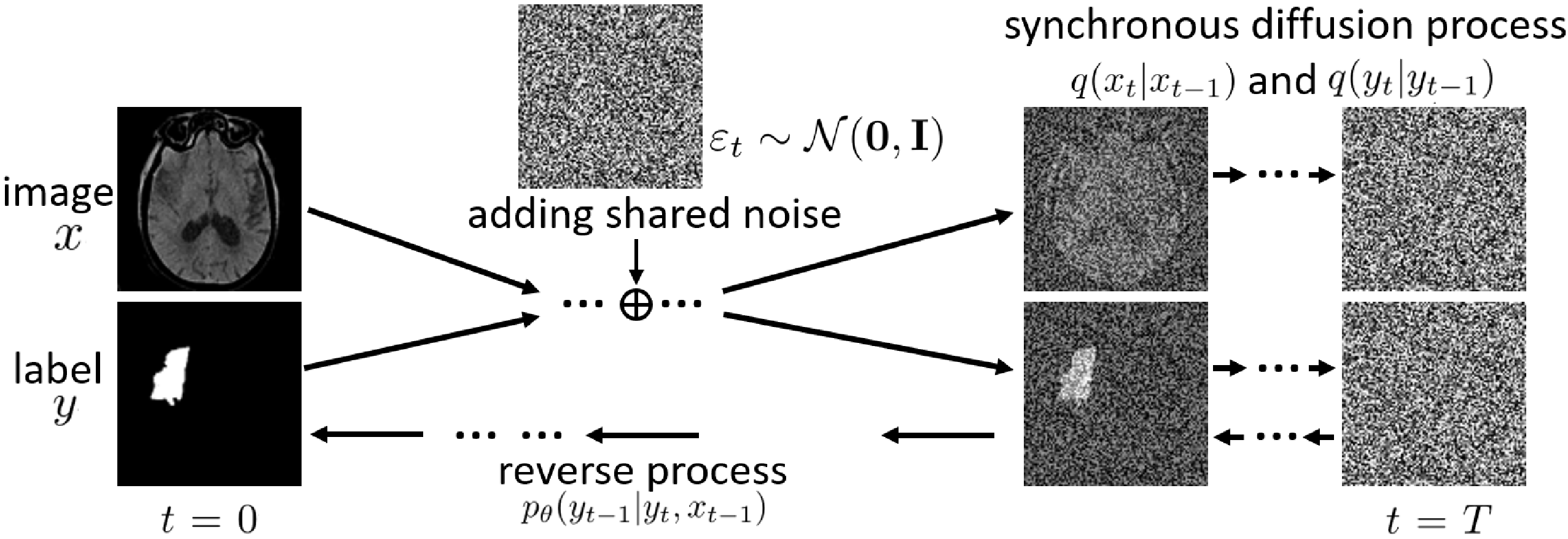}
\caption{Synchronous diffusion and reverse processes utilized in our proposed SDPM}
\label{newfig1}
\vspace{-1em}
\end{figure*}

In this paper, based on a fully generative Latent Variable Model (LVM), a Synchronous image-label Diffusion Probability Model (SDPM) is proposed for efficiently inferring segmentation labels. To this end, we developed a specified variant of variational inference method and a set of related strategies, including synchronous image-label diffusion process illustrated in Fig.1 and a two-stream network of predicting initial noisy label where the inference process starts, to fit our segmentation task, efficiently restoring the segmentation labels from noisy initials. Accordingly, the label inference methods are implemented in four different ways with their own strength, making SDPM more flexible and applicable.
Our contributions briefly include:

\textbf{1)} SDPM is proposed using a synchronous Markov diffusion process for medical image segmentation task. SDPM is based on the fully generative latent variable model with derivational interpretability.

\textbf{2)} A specified variational inference and the involved strategies are proposed for training SDPM. Four inference algorithms in different ways are proposed to efficiently obtain final labels.

\negthinspace
\section{Methodology}
\subsection{Revisit diffusion model}
\negthinspace
As a LVM, diffusion model\cite{sohl2015deep,ho2020denoising} has been successfully applied to the field of image generation with the form $p_\theta(x_0) \triangleq \int p_\theta(x_{0:T})d{x_{1:T}}$, where $x_{1:T}$ are the latent variables,  $x_0$ follows an approximately sampled distribution $q(x_0)$. The joint distribution $p_\theta(x_{0:T})$ is called the reverse process~\cite{sommer2015anisotropic} modeled by a first-order Markov chain starting at $x_T\sim\mathcal{N}(\mathbf{0},\mathbf{I})$.
The diffusion process $q(x_{1:T}|x_0)$ is also a Markov chain, which adds standard normal noise with a variance schedule $\beta_t$. All the mathematical notations are as same as in the paper \cite{ho2020denoising}:
\begin{equation}
p_\theta(x_{0:T})\!\triangleq p(x_T)\!\prod\nolimits_{t=1}^{T}p_\theta(x_{t-1}|x_t), \texttt{   }q(x_{1:T}|x_0)\triangleq \prod\nolimits_{t=1}^{T}q(x_{t}|x_{t-1})
\end{equation}
Minimizing the Kullback-Leibler (KL) Divergence between $q(x_{1:T}|x_{0})$ and $p_\theta( x_{0:T})$ will optimize the variational boundary on the negative log-likelihood. To efficiently implement this minimization, two properties are utilized: 1) the observation $x_t$ is sampled at any time in a closed form:
\begin{equation}
q(x_{t}|x_0)\triangleq \mathcal{N}(x_t|\sqrt{\bar{\alpha}_t}x_0,\gamma_t\mathbf{I}),\texttt{   }\bar{\alpha}_t=\prod\nolimits_{k=1}^t\alpha_k, \alpha_t=1-\beta_t, \gamma_t=1-\bar{\alpha}_t
\end{equation}
2) a vicarious posterior $q(x_{t-1}|,x_t, x_0)$ with condition $x_0$ is used for an intractable posterior $q(x_{t-1}|x_t)$ when training the model:
\begin{eqnarray}
& & q(x_{t-1}|,x_t, x_0)\triangleq \mathcal{N}(x_{t-1}|\tilde{\mu}_t(x_t,x_0),\tilde{\beta}_t\mathbf{I})\nonumber\\
\texttt{where} & & \tilde{\mu}_t(x_t,x_0)={\sqrt{\bar{\alpha}_{t-1}}\beta_t\gamma_t^{-1}}x_0+{\sqrt{{\alpha}_t}\gamma_{t-1}\gamma_t^{-1}}x_t\texttt{ and }\tilde{\beta}_t={\beta_t\gamma_{t-1}\gamma_t^{-1}}
\end{eqnarray}
After simplifying the loss function, the optimization process is equivalent to predict the noise $\varepsilon_t$ from the diffused image $x_t$ using a neural network:
\begin{equation}
\mathcal{L}=\mathbb{E}_{x_t,\varepsilon_t\sim\mathcal{N}(\mathbf{0},\mathbf{I})}\big[{2\sigma^2_t{\alpha}_t{\beta}_{t}^{-2}\gamma_t}\Vert\varepsilon_t-\hat{\varepsilon}_t(x_t,t)\Vert^2\big]
\end{equation}
The convergent model is capable of inferring the unseen images from the random noise following the standard normal distribution.

\subsection{SDPM for Semantic Segmentation}
We extend the diffusion model to segmentation tasks with the form $p_\theta(y_0|x_0)=p_\theta(y_0,x_0)/p_\theta(x_0)\propto p_\theta(y_0,x_0)$, where $p_\theta(y_0,x_0)\triangleq\int p_\theta(x_{0:T}, y_{0:T})dx_{1:T},y_{1:T}$, and $y_{1:T}$ are new members of latent variables. In the new model, the joint distribution $p_\theta(x_{0:T}, y_{0:T})$ is defined as the reverse process, and it can be factorized as an original DDPM~\cite{ho2020denoising} part $p_\theta(x_{0:T})$ and a conditional reverse process $p_\theta( y_{0:T}|x_{0:T})$, which is a series of terms by Markov chains starting at $p_\theta(y_T|x_T)$:
\begin{equation}
p_\theta( y_{0:T}|x_{0:T})=p_\theta(y_T|x_T)\prod\nolimits_{t=1}^T p_\theta( y_{t-1}|y_t,x_{t-1})
\end{equation}
The diffusion process is an approximate conditional posterior $q(y_{1:T}|y_0,x_{0:T})$: 
\begin{equation}
q_( y_{1:T}|y_0,x_{0:T})=\prod\nolimits_{t=1}^T q( y_{t}|y_{t-1},x_{t})
\end{equation}
\negthinspace
A KL Divergence is defined between the diffusion process $q_( y_{1:T}|y_0,x_{0:T})$ and the reverse process $p_\theta( y_{0:T}|x_{0:T})$ to obtain the variational upper bound on the negative log likelihood because of the non-negative property of KL divergence~\cite{lopez2018information}:
\negthinspace
\begin{equation}
\mathbb{KL}\big(q(y_{1:T}|y_0,x_{0:T})\Vert p_\theta(y_{1:T}|y_0,x_{0:T})\big)-\mathbb{E}_q[\texttt{log}p_\theta(y_0|x_0)] \geqslant 0
\end{equation}
The loss function is then defined as minimizing the KL divergence of the conditional posterior and the unnormalized distribution:
\negthinspace \negthinspace
\begin{eqnarray}
\!\!\!\!\!\mathcal{L}&\!\!\triangleq& \!\mathbb{E}_q\Big[\texttt{log}\frac{p_\theta(y_0|x_0)}{p_\theta(y_T|x_T)}+\sum\nolimits_{t=1}^T\texttt{log}\frac{q(y_t|y_{t-1},x_t)}{p_\theta(y_{t-1}| y_t, x_{t-1})}-\texttt{log}p_\theta(y_0|x_0)\Big]\nonumber\\
\label{eq9}
\!&\!\!=&\!\!\!\mathbb{E}_q\Big[\texttt{log}q(y_T|y_0)+\!\sum_{t=2}^T\texttt{log}\frac{q(y_{t-1}|y_{t},y_0,x_t)}{p_\theta(y_{t-1}| y_t,  x_{t-1})}\Big]-\!\texttt{log}p_\theta(y_0|y_1,x_0)p_\theta(y_T|x_T)
\end{eqnarray}
Eq.(\ref{eq9}) indicates the acquisition of $y_0$ could be the traditional way of outputting the label $x_0\mapsto y_0$, or a generative way of Markov chains by inference $x_T\mapsto y_0$. Namely, as long as an initial label estimate $y_T$ exists, the trained diffusion model can infer back and get the label $y_0$. To this end, an additional sub-network for estimating the noisy label $y_T$ from the image $x_T$ is introduced into the network.

Unfortunately, it is difficult to estimate a proper initial $y_T$ for the subsequent inference, because image information is severely destroyed. It is less likely to predict $y_T$ from the image $x_T$ nearly following the distribution $\mathcal{N}(\mathbf{0},\mathbf{I})$. Thus, $-\texttt{log}p_\theta(y_T|x_T)$ is a very strong restriction for predicting the initial $y_T$. Using the final initial $y_T$ could produce a poor result and degrade the segmentation performance notably. To obtain a good initial $y_t$ for inferring $y_0$, we add a time window of length $T_p$ to train the model at each time (the loss $\mathcal{L}_p$ in Eq.(\ref{e1})), guaranteeing that the label $y_0$ could be efficiently restored.
Introducing the prospect to predict an initial $y_t$, the acquisitions of $y_{t-1}$ do not solely depend on the immediate outputs of the network fed by the noisy image $x_{t-1}$ any longer.
The term $\mathbb{E}_q\big[\texttt{log}q(y_T|y_0)\big]$ is constant and thus can be omitted. Therefore, the loss function is further simplified as:

\negthinspace
\begin{equation}
\label{e1}
\!\mathcal{L}=\!\underbrace{\mathbb{E}_q\Big[\sum\nolimits_{t=2}^T\texttt{log}\frac{q(y_{t-1}|y_{t},y_0,x_t)}{p_\theta(y_{t-1}| y_t)}\Big]}_{\mathcal{L}_{d}}-\underbrace{\sum\nolimits_{t=0}^{T_p}\texttt{log}p_\theta(y_t|x_t)}_{\mathcal{L}_{p}}-\underbrace{\texttt{log}p_\theta(y_0|y_1)}_{\mathcal{L}_{d_0}}
\end{equation}

For the loss $\mathcal{L}_d$, the term $p_\theta(y_{t-1}|y_t)$ in reverse process is still compared using KL divergence by the conditional posterior $q(y_{t-1}|y_t,y_0,x_t)$ in the diffusion process introduced in \cite{ho2020denoising}. To make the distribution $q(y_{t-1}|y_t,y_0,x_t)$ tractable, a further assumption is made that images and labels are both overlain by the same Gasussian noise during the diffusion process, i.e., synchronous image-label diffusion process. Thus, the posterior $q(y_{t-1}|y_{t},y_0,x_t)$ is degenerated as $q(y_{t-1}|y_0,\varepsilon_t)$, which only relies on the noise at time $t$ and $y_0$:
\begin{eqnarray}
& &q(y_{t-1}|y_0,\varepsilon_t) = \mathcal{N}(y_{t-1}|\tilde{\mu}_t, \tilde{\beta}_t\mathbf{I})\nonumber\\
\texttt{where  } & & \tilde{\mu}_t = \sqrt{\bar{\alpha}_{t-1}}y_0+{\sqrt{\alpha_t}({\gamma}_{t-1}\gamma_t^{-\frac{1}{2}})}\varepsilon_t, \texttt{   }\tilde{\beta}_t =\beta_t{\gamma_{t-1}}\gamma_{t}^{-1}
\end{eqnarray}
Since the diffusion process for images and labels are shared with the same noise $\varepsilon_t\sim\mathcal{N}(\mathbf{0}, \mathbf{I})$, the images and labels at time $t$ can thus be sampled as:
\negthinspace
\begin{equation}
\label{e8}
x_t = \sqrt{\bar{\alpha}_t}x_0+\sqrt{\gamma_t}\varepsilon_t,\texttt{   }y_t = \sqrt{\bar{\alpha}_t}y_0+\sqrt{\gamma_t}\varepsilon_t
\end{equation}

\begin{figure*}[t]
\center
\includegraphics[width=0.8\textwidth]{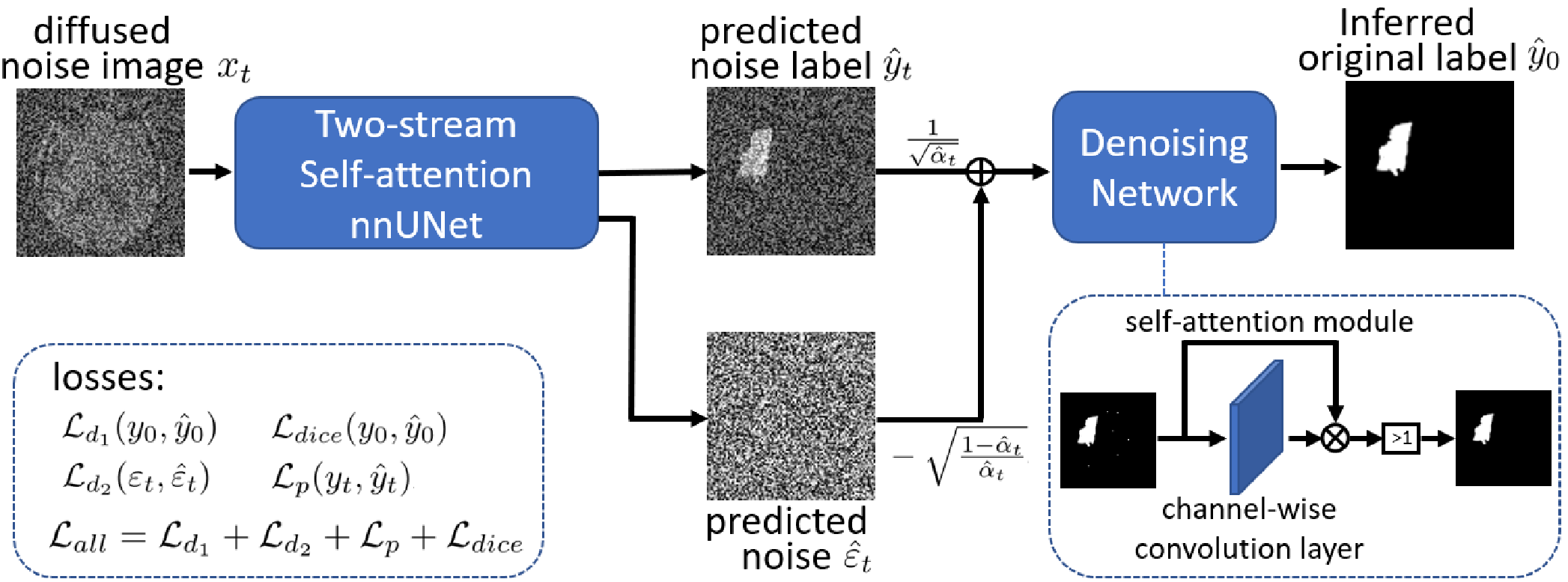}
\caption{The training process of the proposed SDPM with the supervised loss functions}
\label{newfig2}
\vspace{-1em}
\end{figure*}

\negthinspace
The loss $\mathcal{L}_{d_0}$ is for optimizing the last step of generating $y_0$. Some cumbersome discrete segmentation points might degrade the performance to some degree. To refrain from the influence from those points, in addition to applying the same strategy of stopping adding the uncertainty inference of variance $\tilde{\beta}_1$ in \cite{ho2020denoising}, we also append a self-attention module with a convolutional layer of frozen parameters of all ones to get rid of the discrete points with a proper threshold, as illustrated in Fig.\ref{newfig2}.

\negthinspace \negthinspace \negthinspace
\subsection{Optimizing SDPM}
\negthinspace
The choice for the transition term $p_\theta(y_{t-1}|y_t)$ in the reverse process is still a Gaussian distribution, i.e., $\mathcal{N}\big(\mu_\theta(y_t,t), \sigma_t^2\mathbf{I}\big)$, where $\theta$ is the trainable parameters and $\sigma_t^2=\beta_t$ or $\sigma_t^2=\tilde{\beta_t}$ are suggested in \cite{ho2020denoising}. The loss $\mathcal{L}_{d}$ is reparameterized as:
\begin{equation}
\mathcal{L}_d=\mathbb{E}_q\Big[\frac{1}{2\sigma_t^2}\Vert \tilde{\mu}_t(y_0, \varepsilon_t)-\mu_\theta(y_0, \varepsilon_t)\Vert^2\Big]-\underbrace{{2^{-1}D}\texttt{log}\tilde{\beta}+D\texttt{log}\sigma_t}_{C}
\end{equation}
where $D$ is image dimensionality.
The loss function reveals that as long as the model $\mu_\theta$ is able to predict $\tilde{\mu}_t$ given the label $y_0$ with the shared $\varepsilon_t$ in the posterior, the final label $y_0$ is able to be inferred by Markov process.

To further reduce the uncertainty of inferring $y_0$ through stochastic samplings of $q(y_{t-1}|y_0,\varepsilon_t)$, different from the original DDPM~\cite{ho2020denoising}, SDPM adds another loss function to restrict the difference between the true and predicted final labels $y_0$:
\negthinspace
\begin{equation}
\mathcal{L}_d=\mathbb{E}_{y_0,\varepsilon_t\sim\mathcal{N}(\mathbf{0},\mathbf{I})}\Big[\underbrace{\frac{{\bar{\alpha}_{t-1}}}{2\sigma^2_t}\Vert y_0-\hat{y}_0(\hat{y}_t,\hat{\varepsilon}_t,t)\Vert^2}_{\mathcal{L}_{d_1}}+\underbrace{\frac{{{\alpha}_{t}\gamma_{t-1}^2}}{2\sigma^2_t\gamma_t}\Vert\varepsilon_t-\hat{\varepsilon}_t(y_t,t)\Vert^2}_{\mathcal{L}_{d_2}}\Big]+\xi
\end{equation}
where $\hat{y}_0=\hat{y}_t/{\sqrt{\bar{\alpha}_t}}-\sqrt{\gamma_t/\bar{\alpha}_t}\hat{\varepsilon}_t$ and $\xi$ is a binomial residue term. As the losses $\mathcal{L}_{d_1}$ and $\mathcal{L}_{d_2}$ are getting optimized, the $\xi$ is also getting optimized and thus has been ignored for now.
Notice that it is impossible to predict $\varepsilon_t$ from the label $y_t$,

\noindent because there is no image information from $y_t$ at all.
Fortunately, from~\cite{baranchuk2021label,voynov2022sketch}, the network predicting the noise at each diffusion process also carries the label information. Thus, the model can utilize $x_t$ to predict the shared noise $\hat{\varepsilon}_t(x_t,t)$.

For the loss $\mathcal{L}_{p}$, since the network can predict stochastic noise, the noisy labels $y_t$ can thus be predicted with the supervision of the noised label: \negthinspace
\begin{equation}
\mathcal{L}_p=\mathbb{E}_{x_t\sim q(x_{t}|x_0)}\Big[\Vert y_t-\hat{y}_t(x_t,t)\Vert^2\Big]
\end{equation}
To further improve the segmentation performance, a classic dice loss can also be applied to this composite loss function with the treatment of adding an activation function of sigmoid directly at the last layer of outputting $y_0$. The supervised loss functions are illustrated in Fig.\ref{newfig2}.

\negthinspace
\subsection{Inferring the Labels by SDPM}
\negthinspace
The final label $y_0$ can be inferred in four ways. The first is the fast and easy way by directly outputting: $\hat{y}_0^{\mathbf{avg}}=\hat{y}_0$, where $\hat{y}_0$ is the output of the trained network given the clean image $x_0$. The second is based on the salient weight $\psi_t$ over the time window $T_i$: $\hat{y}_0^{\mathbf{sal}}=\frac{1}{N}\sum_{n=1}^{N}\big(\sum_{t=0}^{T_i<T}\hat{y}_0\psi_t\big)$, where $\psi_t=1-(\frac{t}{T_i})^\nu$, $\nu>1$ are gradually degraded coefficients. The third is based on the Markov chain inference starting at time $T_i$: $\hat{y}_0^{\mathbf{infer}}=\frac{1}{N}\sum_{n=1}^{N}\big(\mathbf{IL}(d_i, T_i)\big)$, where $d_i\in\mathbb{R}^{(0,1)}, T_i\in\{0,\cdots,T\}$ and $\mathbf{IL}(\cdot)$ is the Algorithm 1.
The last is the union of all the results: $\hat{y}_0^{\mathbf{all}}=\hat{y}_0^{\mathbf{avg}}\cup\hat{y}_0^{\mathbf{sal}}\cup\hat{y}_0^{\mathbf{infer}}$.
Note that the second and third inferences are performed $N$ times because of the randomness of noise. The average value of $N$ time is used as the final result.

\vspace{-1em}
\begin{algorithm}[!t]
\caption{Inferencing Labels(IL) in Reverse Process}
\begin{algorithmic}[1]
\State \textbf{Input}: image $x_0$, $\texttt{    }$ \textbf{Parameters}: $d_i\in\mathbb{R}^{(0,1)}$, $T_i<T$
\State \textbf{Initialization}:  noised image $\hat{x}_{T_i}=x_{T_i}$ and noised label $\hat{y}_{T_i}$ using (\ref{e8})
\State                noise estimate $\hat\varepsilon_{T_i}$ using trained neural network
\For{$t=T_i,\cdots,1$}
\State estimate $\hat{x}_0$, $\hat{y}_0$ using (\ref{e8})
\If{$t>1$}
\State sampling ${\varepsilon}\sim\mathcal{N}(\mathbf{0},\mathbf{I})$: estimate $\hat{x}_{t-1}=\sqrt{\bar{\alpha}_{t-1}}\hat{x}_0+\frac{\sqrt{\alpha_t}\gamma_{t-1}}{\sqrt{\gamma_t}}\hat\varepsilon_t+\frac{\beta_t\gamma_{t-1}}{\gamma_t}\varepsilon$
\State $\texttt{                    }$ and $\hat{y}_{t-1}=\sqrt{\bar{\alpha}_{t-1}}\hat{y}_0+\frac{\sqrt{\alpha_t}\gamma_{t-1}}{\sqrt{\gamma_t}}\hat\varepsilon_t+\frac{d_i\beta_t\gamma_{t-1}}{\gamma_t}\varepsilon$
\State update next loop noise $\hat\varepsilon_{t-1}$ from $\hat{x}_{t-1}$ using trained neural network
\ElsIf{$t=1$}
\State estimate $\hat{y}_0$ using (\ref{e8}), $\hat{y}_{0}^\mathbf{final}=\sqrt{\bar{\alpha}_{t-1}}\hat{y}_0+\frac{\sqrt{\alpha_t}\gamma_{t-1}}{\sqrt{\gamma_t}}\hat\varepsilon_t$
\EndIf
\EndFor
\State \textbf{Output}: label estimate $\hat{y}_{0}^\mathbf{final}$
\end{algorithmic}
\end{algorithm}

\negthinspace
\section{Experiments and Results}
\negthinspace
\noindent\textbf{Datasets and Pre-cessing}: Two datasets were involved:
1) A private dataset, named \textbf{Infarct}, containing 195 AIS patient NCCT scans (5 mm) were included. Of 195 patients, 123 images were used for training while the remained 72 for testing.
2) Another private dataset comprising of 331 patients with acute intracranial hemorrhage confirmed by NCCT (2.5mm), called \textbf{Hemorrhage}, was also included patients with acute ICH confirmed by NCCT (2.5 mm thickness). Of 331 patient scans, 241 scans were used for training and validation, and 90 were used for testing.

\noindent\textbf{Hyper-parameters Settings:}
Online data augmentations was performed, including adding noise, rotations, scalings, inplane flipping, etc. Learning rate was reduced non-linearly from 1e-4 to 6e-5 with the Adam optimizer, where $lr=lr_{init}*(1-i_c/i_{max})^{0.9}+lr_{min}*(i_c/i_{max})^{0.9}$, $i_c$ is the current iteration and $i_{max}$ is the maximum number of iteration. The variance schedule $\beta_t$ is the sigmoid curve\cite{nichol2021improved}. P2 Weighting coefficients during the training suggested in \cite{choi2022perception}.
The repeated times are 100 and 50 for $\hat{y}_0^{\mathbf{sal}}$ and $\hat{y}_0^{\mathbf{infer}}$ in the inference process. Diffusion period $T$ is 500 and the initial time $T_i=T/2$.

\noindent\textbf{Evaluation Metrics:}
Three metrics\cite{taha2015metrics} including Dice, Volume Correlation(VC) based on the Pearson product-moment correlation coefficient, and Volume Difference Percentages(VDP), were used to quantitatively assess the performance of the model prediction at a voxel level compared to manual contouring.
Five other methods, including SegResNet, UNETR, SwinUNETR, nnUNet, nnUnet++, were also applied on the same three datasets.
For fair comparisons, the best performance with the optimized parameters for each method was reported.

\begin{figure*}[t]
\center
\includegraphics[width=0.95\textwidth]{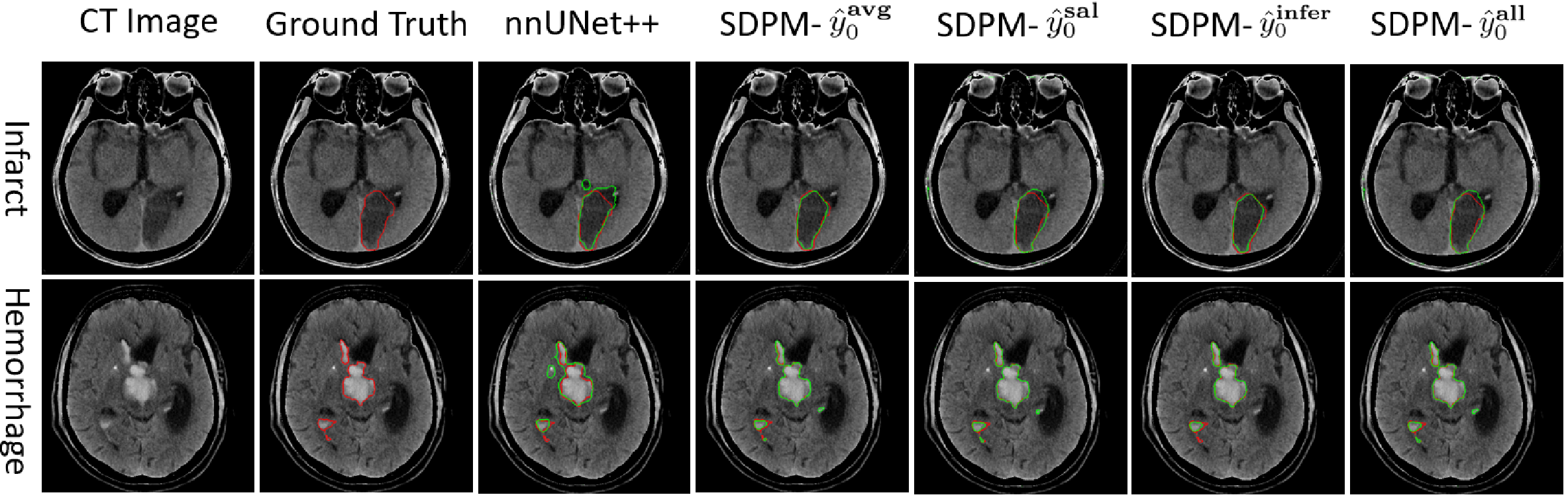}
\caption{Qualitative results for infarct and hemorrhage segmentations on two datasets}
\label{newfig3}
\vspace{-1.5em}
\end{figure*}

\noindent \textbf{Results:} A few segmentation examples are visualized in Fig.\ref{newfig3}. Quantitative results in Table.\ref{tab1} show SDPM with four inferences obtained the best performance on the whole with Dice, VC and VDP.
The method 'CDPM w/o noise' is with the same neural network architecture but without diffusion process.
In our private Infarct dataset, although SDPM w/o noise and $\hat{y}_0^{\mathbf{all}}$ have obtained nearly same dice of 0.4, SDPM-$\hat{y}_0^{\mathbf{all}}$ have better VC-0.619 and lower VDP-0.545.
In hemorrhage dataset, the best performance by $\hat{y}_0^{\mathbf{infer}}$  is Dice-0.931, VC-0.985 and VDP-0.032.
Additionally, Table.\ref{tab1} also suggested that our method was able to greatly reduce the VDP scores across three datasets.

\negthinspace
\begin{table}
\centering
\caption{Quantitative performance on \textbf{Infarct} and \textbf{Hemorrhage}.}
\label{tab1}
\begin{tabular}{|c|c|c|c|c|c|c|}
\hline
 Datasets  &\multicolumn{3}{c|}{\textbf{Infarct}}&\multicolumn{3}{c|}{\textbf{Hemorrhage}}\\
\hline
Methods  & Dice &  VC & VDP & Dice &  VC & VDP  \\
\hline
SegResNet\cite{myronenko20193d}        & 0.334 & 0.409 & 0.574 & 0.926 & 0.976 & 0.086 \\
UNETR\cite{hatamizadeh2022unetr}       & 0.304 & 0.392 & 0.709 & 0.910 & 0.967 & 0.100\\
SwinUNETR\cite{hatamizadeh2022swin}    & 0.366 & 0.455 & 0.656 & 0.908 & 0.965 & 0.101\\
nnUNet\cite{isensee2021nnu}            & 0.337 & 0.488 & 0.692 & 0.922 & 0.978 & 0.096\\
nnUNet++\cite{zhou2018unet++}          & 0.376 & 0.507 & 0.639 & 0.915 & 0.968 & 0.070\\
SegDiff\cite{amit2021segdiff}          & 0.389 & 0.482 & 0.658 & 0.927 & 0.976 & 0.066\\
MedSegDiff\cite{wu2022medsegdiff}      & 0.396 & 0.540 & 0.657 & 0.930 & 0.976 & 0.069\\
\hline
SDPM w/o noise\cite{ho2020denoising}  & \textbf{0.400} & 0.557 & 0.633 & 0.927 & 0.961 & 0.074\\
\hline
SDPM-$\hat{y}_0^{\mathbf{avg}}$  & 0.378 & 0.544 & 0.641 & 0.929 & 0.972 & 0.078\\
SDPM-$\hat{y}_0^{\mathbf{infer}}$  & 0.374 & 0.531 & 0.652 & \textbf{0.931} & \textbf{0.985} & 0.067\\
SDPM-$\hat{y}_0^{\mathbf{sal}}$  & 0.397 & 0.581 & \textbf{0.540} & 0.928 & 0.980 & \textbf{0.032}\\
\hline
SDPM-$\hat{y}_0^{\mathbf{all}}$ & 0.399 & \textbf{0.619} & 0.545 & 0.928 & 0.980 & \textbf{0.032}\\
\hline
\end{tabular}
\vspace{-0.6em}
\end{table}

\negthinspace \negthinspace \negthinspace \negthinspace
\section{Discussion and Conclusion}
\negthinspace \negthinspace \negthinspace \negthinspace
A novel probabilistic SDPM is proposed to automatically segment stroke lesions of hemorrhage and infarct on NCCT, in order to alleviate the tedious manual contouring currently used in clinic~\cite{samuels2023infarct}.
The segmentation labels are output by SDPM in a fully probabilistic generative way. With the proposed several inference methods, the model was able to efficiently recover the lesion labels.
Compared to the reference standard of manual contouring, quantitative evaluations demonstrate the efficacy of the proposed SDPM, outperforming several CNN and transformer based methods.

This study  represents the first study to use a completely probabilistic inference model based on DPM to automatically segment infarct and hemorrhage on NCCT. Table.\ref{tab1} has shown the proposed SDPM with four inference methods obtained the start-of-the-art performance on two datasets. All the ablation studies in Table.\ref{tab1} revealed every inference method has its own strength, where salient weighting estimate $\hat{y}_0^{\mathbf{all}}$ reached the best VDP of 0.540 and 0.032, $\hat{y}_0^{\mathbf{sal}}$ obtained the best Dice in two infarct datasets. In the hemorrhage dataset, inference method $\hat{y}_0^{\mathbf{infer}}$ reached the best results of Dice and VC. Generally, the inference of immediately outputting the labels performed worse than other three inference methods.

This study has several limitations. First, our datasets are limited. More training samples may further improve the segmentation accuracy and generalizability. Second, the final label inference is slightly influenced by stochastic factors. It is time-consuming to get an average prediction based on several inferences. Third, the predicted labels generated by different inference methods were simply averaged. More advanced label fusion techniques may improve the performance.

In conclusion, a synchronous image-label diffusion model by a LVM is proposed to segment stroke lesion of infarct and hemorrhage on NCCT. Experiments on three datasets demonstrate the efficacy of the proposed method, suggesting its potentials of being used for stroke lesion volume measurement.

 \bibliographystyle{IEEEtran}
 \bibliography{mybibliography}
%
%
%
%
%
\end{document}